\RequirePackage{fix-cm}
\documentclass[smallextended]{svjour3}       
\smartqed  
\usepackage{graphicx}
\usepackage[caption=false]{subfig}
\usepackage[numbers]{natbib}
\usepackage{float}
\usepackage{multirow}

\journalname{IJCARS}
\begin{document}

\title{Learning Deep Similarity Metric for 3D MR-TRUS Registration
\thanks{This work was supported by the Intramural Research Program of the National Institutes of Health, the National Institutes of Health Center for Interventional Oncology, and NIH grants 1ZIDBC011242 and 1ZIDCL040015.}
}

\titlerunning{Learning Deep Similarity Metric for 3D MR-TRUS Registration}        

\author{Grant Haskins \and Jochen Kruecker \and Uwe~Kruger \and Sheng Xu \and Peter~A.~Pinto \and Brad~J.~Wood \and Pingkun~Yan
}

\authorrunning{Haskins et al.} 

\institute{G. Haskins, U. Kruger, P. Yan* \at
              Department of Biomedical Engineering, Rensselaer Polytechnic Institute, Troy, NY 12180, USA\\
              Asterisk indicates corresponding author\\
              Tel.: +1-518-276-4476\\
              \email{yanp2@rpi.edu}           
		\and
		J. Kruecker \at
		Philips Research North America, Cambridge, MA 02141, USA
           \and
           S. Xu, P.A. Pinto, B.J. Wood \at
              National Institutes of Health, Center for Interventional Oncology, Radiology \& Imaging Sciences, Bethesda, MD 20892, USA
}

\date{Received: date / Accepted: date}

\newcommand{\absdiv}[1]{%
  \par\addvspace{.5\baselineskip}
  \noindent\textbf{#1}\quad\ignorespaces
}

\maketitle

\begin{abstract}
\section*{}
\absdiv{Purpose}
The fusion of transrectal ultrasound (TRUS) and magnetic resonance (MR) images for guiding targeted prostate biopsy has significantly improved the biopsy yield of aggressive cancers. A key component of MR-TRUS fusion is image registration. However, it is very challenging to obtain a robust automatic MR-TRUS registration due to the large appearance difference between the two imaging modalities. The work presented in this paper aims to tackle this problem by addressing two challenges: (i) the definition of a suitable similarity metric and (ii) the determination of a suitable optimization strategy.
\absdiv{Methods}
This work proposes the use of a deep convolutional neural network to learn a similarity metric for MR-TRUS registration. We also use a composite optimization strategy that explores the solution space in order to search for a suitable initialization for the second-order optimization of the learned metric. Further, a multi-pass approach is used in order to smooth the metric for optimization.
\absdiv{Results}
The learned similarity metric outperforms the classical mutual information and also the state-of-the-art MIND feature based methods. The results indicate that the overall registration framework has a large capture range. The proposed deep similarity metric based approach obtained a mean TRE of 3.86mm (with an initial TRE of 16mm) for this challenging problem.
\absdiv{Conclusion}
A similarity metric that is learned using a deep neural network can be used to assess the quality of any given image registration and can be used in conjunction with the aforementioned optimization framework to perform automatic registration that is robust to poor initialization. 
\end{abstract}


\section{Introduction}
\label{intro}

Prostate cancer is among the main causes of cancer death in men in the United States \cite{siegel_cancer_nodate}. Although transrectal ultrasound (TRUS) usually has low sensitivity with respect to prostate cancer, it is still the most commonly used imaging modality for guiding prostate biopsy. On the other hand, multi-parametric magnetic resonance (MR) imaging has been shown to have good sensitivity and specificity for identifying a prostate cancer lesion. This is an expensive and time-consuming procedure. Over the past decade, studies have shown that the fusion of TRUS and MR images for guiding prostate biopsies for cancer diagnosis provides clinical benefit by limiting the rate of false negative prostate cancer diagnoses \cite{pinto_magnetic_2011,siddiqui_comparison_2015}. 

Image registration is a key component for multimodal image fusion, which generally refers to the process by which two or more image volumes and their corresponding features (acquired from different sensors, points of view, imaging modalities, etc.) are aligned into the same coordinate space. 
Medical images that are acquired from different imaging modalities use different imaging physics, which creates unique advantages and disadvantages. Relatively unique information about the imaged volume is provided by each modality. Image fusion through registration can integrate the complementary information from multimodal images to help achieve more accurate diagnosis and treatment \cite{reena_benjamin_improved_2018}. In the case of MR-TRUS fusion, the real-time imaging and cost-effective properties of TRUS can be well complemented by the high prostate cancer identification accuracy of MR imaging for image-guided prostate interventions \cite{Fedorov2015}.
Mutual information is the most common pixel-based similarity metric for multi-modality image registration and utilizes the statistical information associated with the image volumes obtained from different modalities \cite{maes1997multimodality,wells1996multi}. However, the correspondence between the alignment with maximum mutual information and the expert alignment for difficult multimodal registration tasks is typically poor because of the inadequate description of the image alignment using pixel intensity mapping. Due to the difficulties associated with directly registering TRUS and MR images, this registration is commonly performed using surface-based methods through shape modeling and the use of feature descriptors \cite{khallaghi_biomechanically_2015,zettinig_multimodal_2015,HeinrichMIND,poulin_validation_2018,fuerst_automatic_2014}. For example, Sun et~al. \cite{sun_efficient_2013} used the modality independent neighborhood descriptor (MIND) \cite{HeinrichMIND} to map the voxels that constitute the MR and TRUS volumes to a descriptor value for comparison. In their image registration framework, the sum of squared differences between the MIND descriptors at the corresponding locations in MR and TRUS images is used as the similarity metric. Although they were able to obtain good registration results in many cases, the quantification of similarity was done using manually crafted feature mapping and can limit the registration performance when the initialization is far from the underlying registration. Sparks et~at. \cite{sparks_fully_2013} developed a fully automatic registration approach utilizing image segmentations to address the appearance difference between the two modalities \cite{sparks_fully_2013}. However, such registration techniques cannot guarantee adequate voxel-to-voxel correspondence of internal structures because these approaches are primarily influenced by the information that is extracted from voxels proximal to the boundary of the prostate.
Unlike the approaches described in the works discussed above, our method uses raw pixel data as its input and uses learned features for estimating the image similarity.

Recently, several works have used deep neural networks to learn application-specific similarity metrics for image registration tasks \cite{cheng_deep_2016,simonovsky_deep_2016,Zagoruyko_Komodakis2015}. Although these existing similarity metric learning methods outperform mutual information and other manually defined metrics for their applications, 
%
the existing deep learning based methods deal with multimodal images that share largely similar views or relatively simple intensity mappings (for example MR-CT or T1-T2 weighted MR images). In our application, the prostate in MR and TRUS looks very different in terms of not only image intensities but also fields of view, which is a much more challenging problem.
In this paper, we propose a deep learning based approach directly registering the two imaging modalities using image pixel intensities. Our primary contributions are two-fold. First, we propose designing a CNN with a skip connection to learn the target registration error (TRE) between 3D MR and TRUS images to act as image similarity metric for registration. Second, we propose a differential evolution initialized Newton-based optimization (DINO) method to perform the optimization and expand the capture range of the registration. To efficiently explore the solution space and also enhance the capture range, the proposed approach uses differential evolution, followed by the local second order algorithm - BFGS. 
The application of the developed approach shows promising performance and also advantages over the classical mutual information and MIND based approaches \cite{sun_three-dimensional_2015}. 

The rest of the paper is organized as follows. 
Section~\ref{sec:methods} presents the details of the proposed method. Section~\ref{sec:training} describes how the convolutional neural network used for learning the similarity metric is trained. The experimental results are given in Section~\ref{sec:experiments}. Finally, Section~\ref{sec:conclusions} draws the conclusions and briefly discusses our future work.

\section{Methods}
\label{sec:methods}

In this section, details of the proposed method are presented. In our application, the transformation is limited to be rigid. Thus, there are three translation and three rotation parameters $\theta_R=\{T_x, T_y, T_z, r_x, r_y, r_z\}$ in the transformation matrix $R$ to be optimized. The reason that we chose to stay with rigid transformation is because it is the registration strategy that is most commonly used in clinical practice, which has also been shown able to obtain clinically significant results \cite{calio_changes_2017,siddiqui_comparison_2015}. 
The rest of this section first briefly discusses convolutional neural networks (CNNs) to introduce some terminologies used in our proposed method. Then the proposed deep similarity metric learning method is presented. Finally, the optimization method used for learning the similarity metric is discussed.

\subsection{Deep Convolutional Neural Networks}

A deep CNN is a neural network that stacks multiple different layers of neurons. CNN has been shown to be very useful for image analysis because of its parameter sharing nature and ability to model local neighborhoods in images. A CNN with a classical architecture typically consists of six types of layers: input layers, convolutional layers, activation function layers, batch normalization (BN) layers, pooling layers, and fully-connected (FC) layers \cite{krizhevsky_imagenet_2012,lecun_deep_2015}. 
The input layer consists of the inputted image volumes, as suggested by its name. The convolutional layers extract feature representations from images by computing the inner product between all of the filters and image patches. Each convolutional layer is usually followed by an activation function layer, which applies an element-wise activation function to the outputted feature map associated with the convolutional layer. A popular activation function is the rectified linear unit (ReLU), which is used in this work and can be expressed as
\begin{equation}
f(x) = \left\{
            \begin{array}{ll}
            x, \textrm{ if } x \geq 0, \\
            0, \textrm{ otherwise.}
            \end{array}
        \right.
\end{equation}
The BN layers, which are also used in this work, reduce the amount of covariance shift to help the CNN adapt to different input intensity scales.
The pooling layer is periodically inserted into a CNN to reduce the size of the feature maps, which not only lowers memory usage but also enlarges the receptive field of a network. The two most common types of pooling layers use max pooling and average pooling. FC layers connect neurons in one layer to every neuron in another layer and are usually used as the last few layers of network. An FC layer can give the CNN a ``global view'' of all the activations but also increase the risk of over fitting.

\subsection{Deep Similarity Metric}

\begin{figure}[tbh]
 \centering
\includegraphics[width=\textwidth]{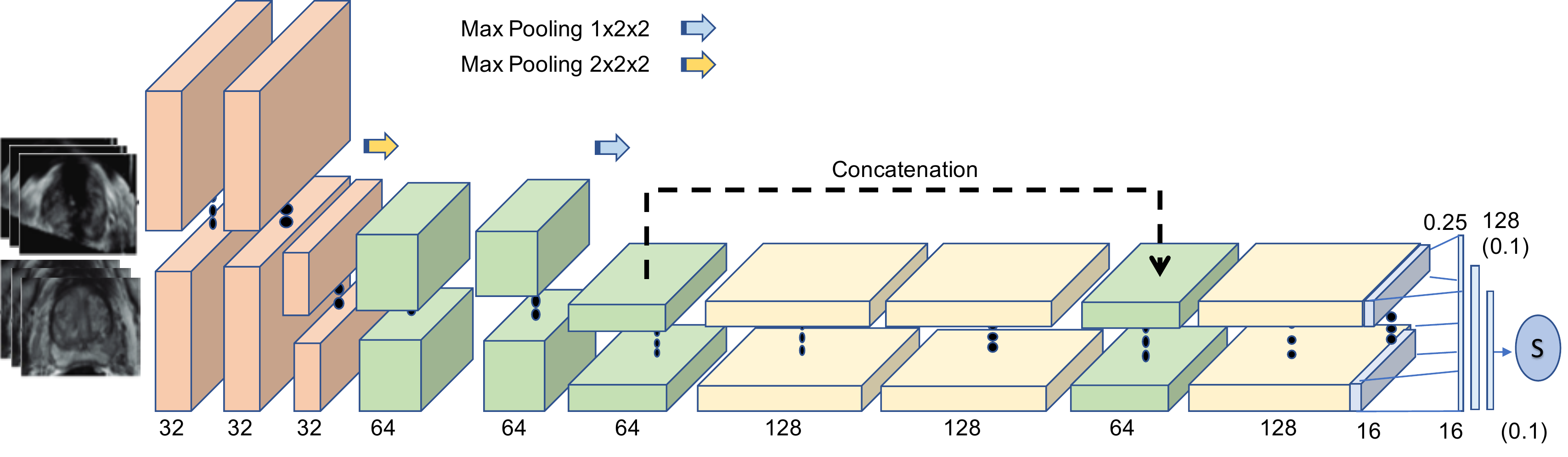} 
\caption{The architecture of the designed CNN that is used to learn the similarity metric.} 
\label{fig:cnn}
\end{figure}

In our proposed method, the determination of the similarity between MR and TRUS images is cast as a deep CNN based regression problem. The CNN takes a pair of 3D MR and TRUS images as its input and outputs an estimate of the target registration error, which is used to assess the quality of an alignment. A 3D MR and TRUS image pair together constitute a single 2-channel image with each image in one channel. A 2-channel input is used because it outperforms other structures as indicated by \cite{simonovsky_deep_2016,Zagoruyko_Komodakis2015}. 
The designed CNN consists of 9 volumetric convolutional layers that use a stride of 1 in each direction. Element-wise Relu activations are used at each layer in order to introduce non-linearity to the model.
A BN layer is used after the second convolutional layer and the concatenation in order to allow the model to be robust to different intensity scales. 
Furthermore, a skip connection is used to catenate the feature maps that are obtained at lower and higher levels in the network. The final fully connected layer outputs a scalar that is used to estimate the target registration error that assesses the quality of the registration. Fig.~\ref{fig:cnn} shows a diagram of the feed forward CNN architecture described above.

\subsection{Image Registration}

Intensity-based image registration that is performed iteratively can be defined by its choice of similarity metric, parameterizable transformation, interpolation strategy, and optimization strategy. Following initialization, the parameters that constitute the affine transformation of the moving image (TRUS) are updated iteratively according to the selected optimization strategy. Since the learned metric is nonconvex and nonsmooth, it is important to design a suitable optimization strategy to enhance the capture range. Many commonly used optimization methods could produce bad registrations for these difficult applications if the moving image is not initialized in such a way that it is sufficiently close to the fixed image.
The resulting affine transformation is then applied to map the moving image to the fixed image at each iteration. Because this mapping rarely results in direct pixel-to-pixel comparisons, robust interpolation (which is linear in this work) is necessary to establish the comparison (defined by the similarity metric) of the resulting two sets of pixel values that are associated with the moving and fixed images. In this paper, we plug the learned similarity metric into the classical image registration framework described above.

Because the designed CNN performs regression for each pair of images independently, the learned similarity metric can be non-smooth and non-convex. In order to address this, we use a multi-pass approach in our registration framework. Throughout the optimization that is used to perform the registration, the moving image is slightly perturbed $N$ times and the average of the associated TRE estimates is used as the objective function evaluation, defined as 
\begin{equation}
E(I_{moving}, I_{fixed}) = \frac{1}{N}\sum_{n=1}^{N} CNN(g(I_{moving}, \theta_n), I_{fixed}),
\label{mpass}
\end{equation}

where $g(\cdot)$ is a resampling function to resample the moving image by using the giving parameter $\theta_n$.
The perturbation parameter $theta_n$ consisting of both rotation and translation is uniformly sampled from the range of [-0.25mm, 0.25mm] for translation and [-1$^{\circ}$, 1$^{\circ}$] for rotation. In our implementation, we used $N=5$. Larger $N$ may lead to smoother curves but will result in higher computational cost.

In our work, to efficiently perform the optimization and expand the capture range of the registration method, we propose a differential evolution initialized Newton-based optimization (DINO) method.
In the proposed method of DINO, differential evolution with early termination is first applied given an initialization. Then the result is used as the initial solution estimate for the Newton-based optimization strategy - the second-order Broyden-Fletcher-Goldfarb-Shanno (BFGS), which approximates the Hessian that is used in typical Newton iterations by using low-rank updates.

We chose to utilize differential evolution because of its demonstrated efficiency in computationally intensive domains and its efficacy dealing with multimodal objective functions \cite{Storn1997}. Differential evolution falls within a class of combinatorial, non-gradient based optimization strategies that is inspired by biological evolution and evolves a population of solutions as opposed to a single solution. It takes a set of possible solutions that purposefully span the solution space and uses random members of the population to evolve the population according to its mutation rate and recombination frequency. ``Children'' only replace ``parents'' if their objective function value - ``fitness'' - is closer to the global minimum. After the termination criteria is met, the most ``fit'' member of the final generation is selected. The algorithm implemented in the SciPy package \cite{scipy} is used in our work and a maximum number of iterations/generations is adopted as the termination criterion.

\begin{figure}[tbh]
 \centering
  \subfloat[]{\includegraphics[width=0.33\textwidth]{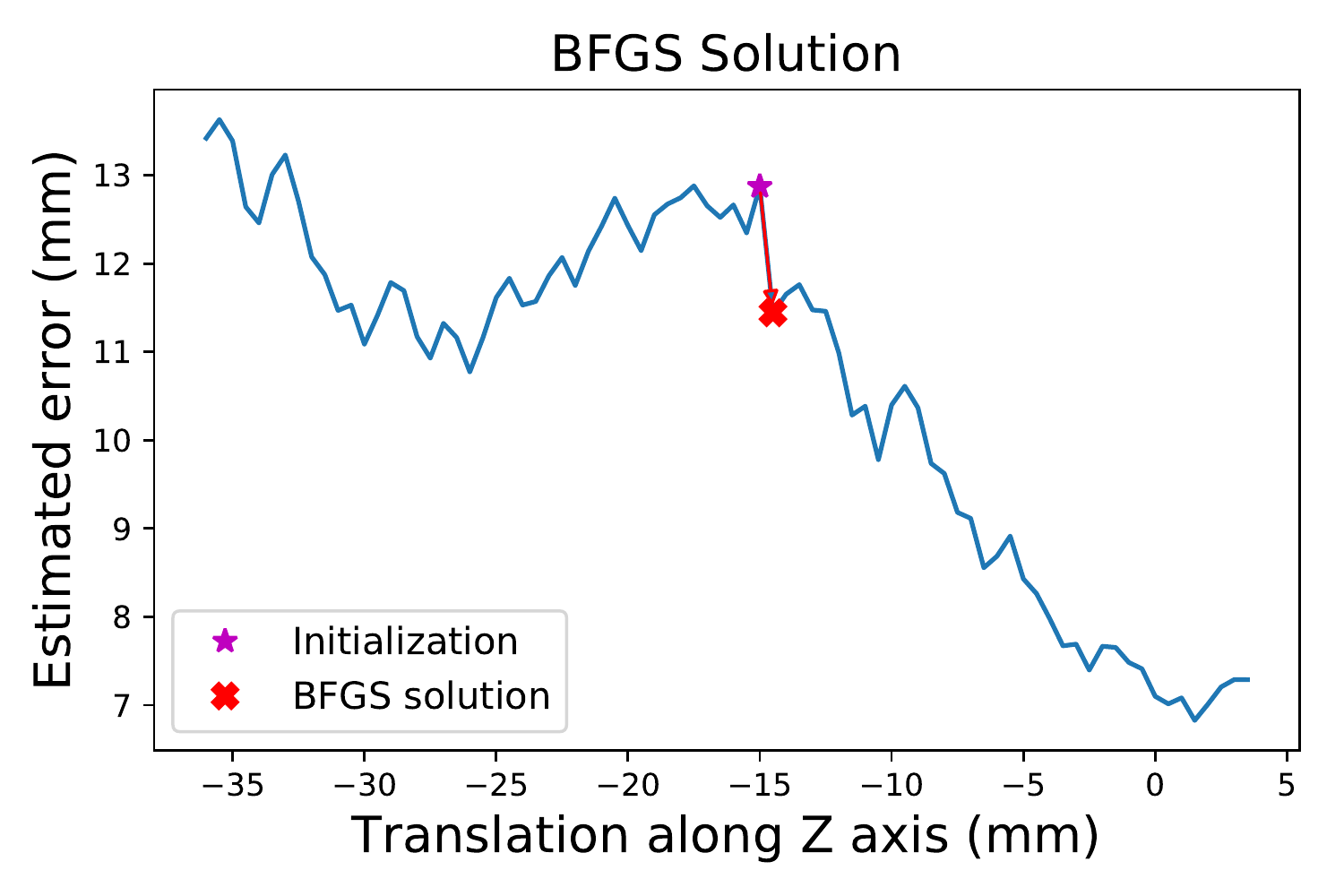}}
  \subfloat[]{\includegraphics[width=0.33\textwidth]{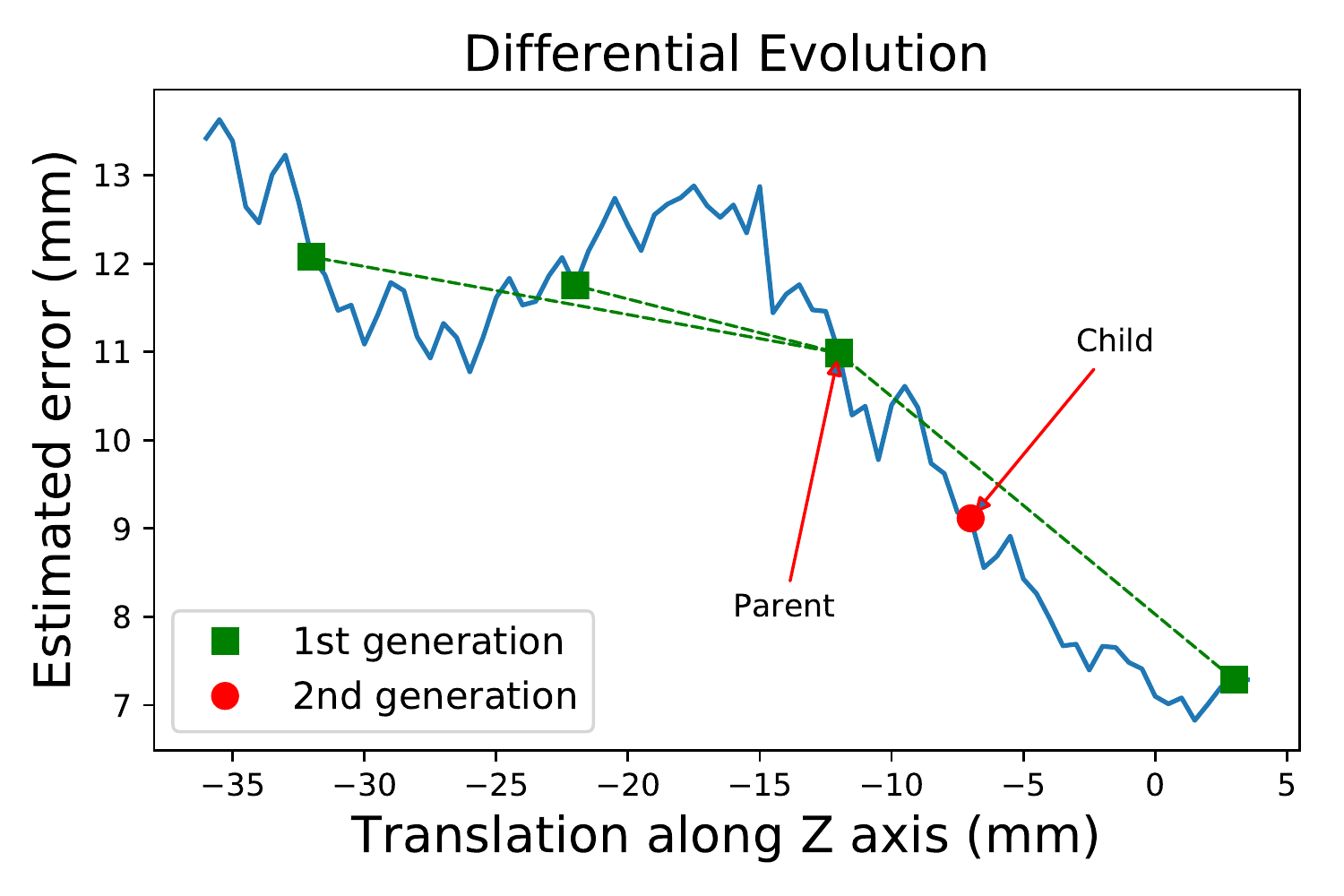}}
  \subfloat[]{\includegraphics[width=0.33\textwidth]{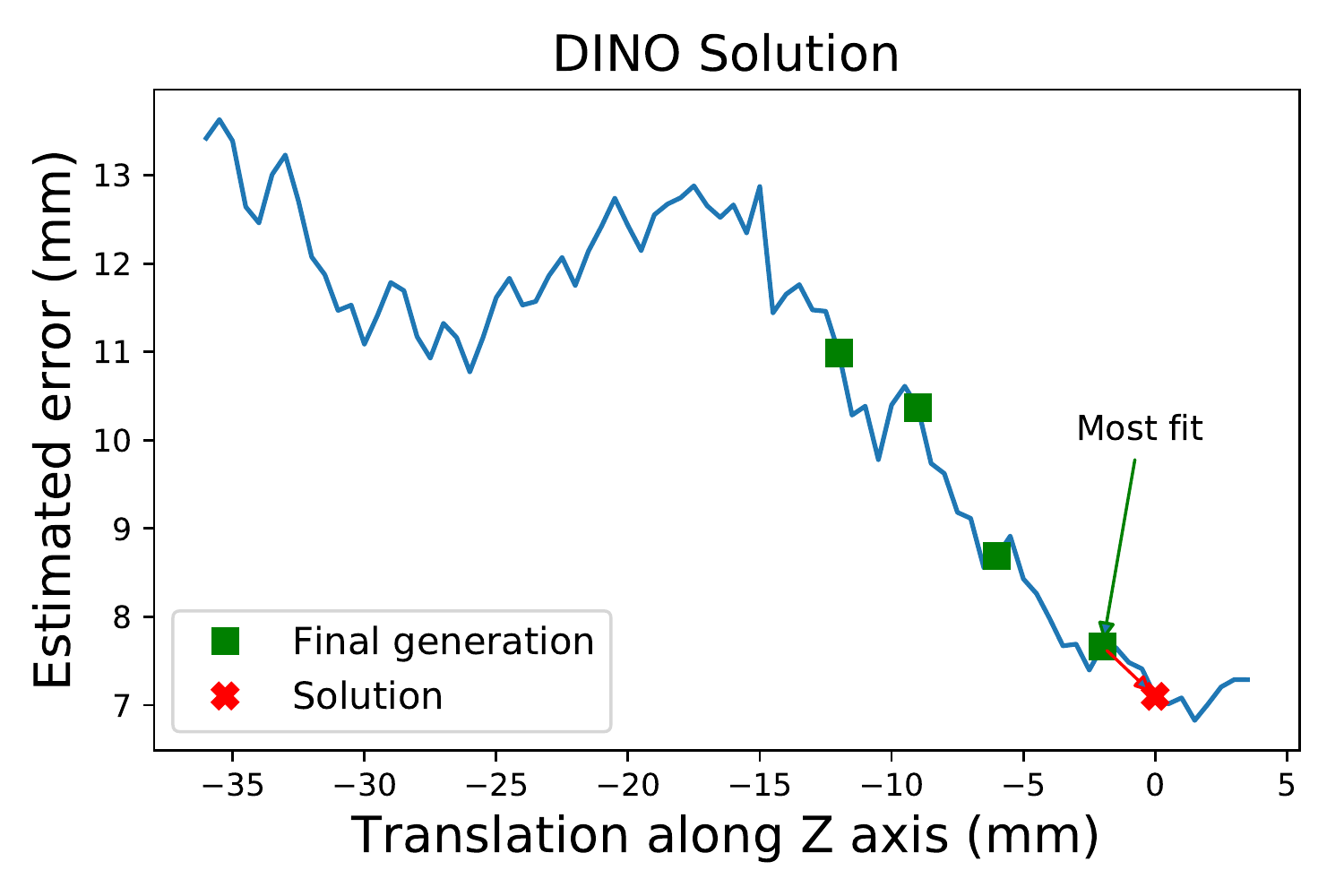}}

\caption{(a) The solution obtained using BFGS to optimize the learned metric using an initial TRE of 16mm. (b) Visualization of the process through which a parent in the first generation of differential evolution produces a child in the second generation. The parent incorporates information from 3 other randomly selected members of the population within its generation to produce the child. This is repeated for every member of the population within the generation. An initial TRE of 16mm is used to produce the first generation, which deliberately spans the space according to the search radius. (c) The solution obtained using the most ``fit'' member of the population of the final generation of differential evolution (member with the lowest learned metric value) to initialize the BFGS algorithm.}       

\label{fig:DINO_plot}
\end{figure}

After truncated differential evolution is used to explore the solution space, the Newton-based optimization strategy (BFGS) that exhibits local, super linear convergence is used to determine the solution. The search radius that is associated with truncated differential evolution that is used in DINO is equal to 20mm (which is the maximum TRE that is used for training). The initial estimate that BFGS requires is provided by the output of the truncated differential evolution algorithm. This overall method (DINO) allows us to explore the solution space in order to expand our capture range. Differential evolution explored the solution space using a population size of 5 for 13 generations before termination and the resulting solution was used as the initialization for BFGS. Further, the ``polish'' argument was set to ``False'' in order to allow us to customize the optimization algorithm used after Differential Evolution. All other parameters for the optimizers were set to their defaults as indicated in Scipy \cite{scipy}. A visualization of the convergence that is realized using both BFGS alone and DINO in order to optimize the learned similarity metric is given in Fig.~\ref{fig:DINO_plot}. Note that the metric is not able to predict the TRE very well when the magnitude of the translation along the z-axis is larger than the maximum TRE associated with the training data.

\section{Network Training and Assessment}
\label{sec:training}

Keras with TensorFlow as the backend \cite{keras} was the library that was used to construct the CNN. The network was trained with the popular optimizer ADAM with a learning rate of 1e-5 using the mean squared error between the regressed TREs and the ground truth TREs as the loss function. Since it is quite difficult to assess the quality of a given 3D MR-TRUS registration, it is important to define robust strategies for generating the samples that are used to train the network and for training the network itself.

\subsection{Training Data Generation}

The network was trained by taking training image pairs that were registered manually by a medical expert who conducted the biopsy procedure and transforming the moving image using known rigid perturbations. Because these perturbations from ground truth are global, we are readily able to obtain the TRE that results from a given perturbation. In our work, TRE is defined as the mean Euclidean distance between the prostate surface points of the prostate in the warped moving image and the corresponding surface points of the prostate in the ground truth moving image. Therefore, fiducial points/landmarks are not involved. The TRE is equal to zero if the warped moving image and the ground truth moving image correspond exactly. As multiple surface points are used, the TRE appropriately reflects the rotatory perturbations. We feed the transformed image pair through our network and use the calculated TRE as the ground truth label in order to compute the loss function (mean squared error). The maximum registration error associated with the known perturbations was 20mm. The transformation parameters are obtained as follows. Random values in either [-3,-1], [-3,-1], [-5,-1], [-12.5,-2.5], [-7.5,-2.5], [-7.5,-2.5] or [1,3], [1,3], [1,5], [2.5,12.5], [2.5,7.5], [2.5,7.5] are sampled uniformly to obtain values for tX, tY, tZ, rotX, rotY, and rotZ, respectively. The smaller parameter values are excluded during training. The parameters are all then scaled according to the magnitude of the TRE associated with the resulting affine transformation matrix.

\subsection{Metric Network Training}

Once the loss function, network architecture, and training data are defined, the next step is to determine the best optimizer to train our neural network. 
In our work, the following optimizers were tested: SGD, SGD with Nesterov momentum, RMSProp, Adagrad, Adam,and Adadelta. Each method used a learning rate of 5$\times 10^{-6}$ and each non-SGD method used the default epsilon values determined by Keras \cite{keras}. RMSProp uses a $\rho$ value of 0.9. Results show that the Adam optimizer (which uses an adaptive learning rate, momentum and an exponentially decaying average) achieved superior performance.

\begin{figure}[tbh]
\centering
  \subfloat[]{\includegraphics[width=0.5\textwidth]{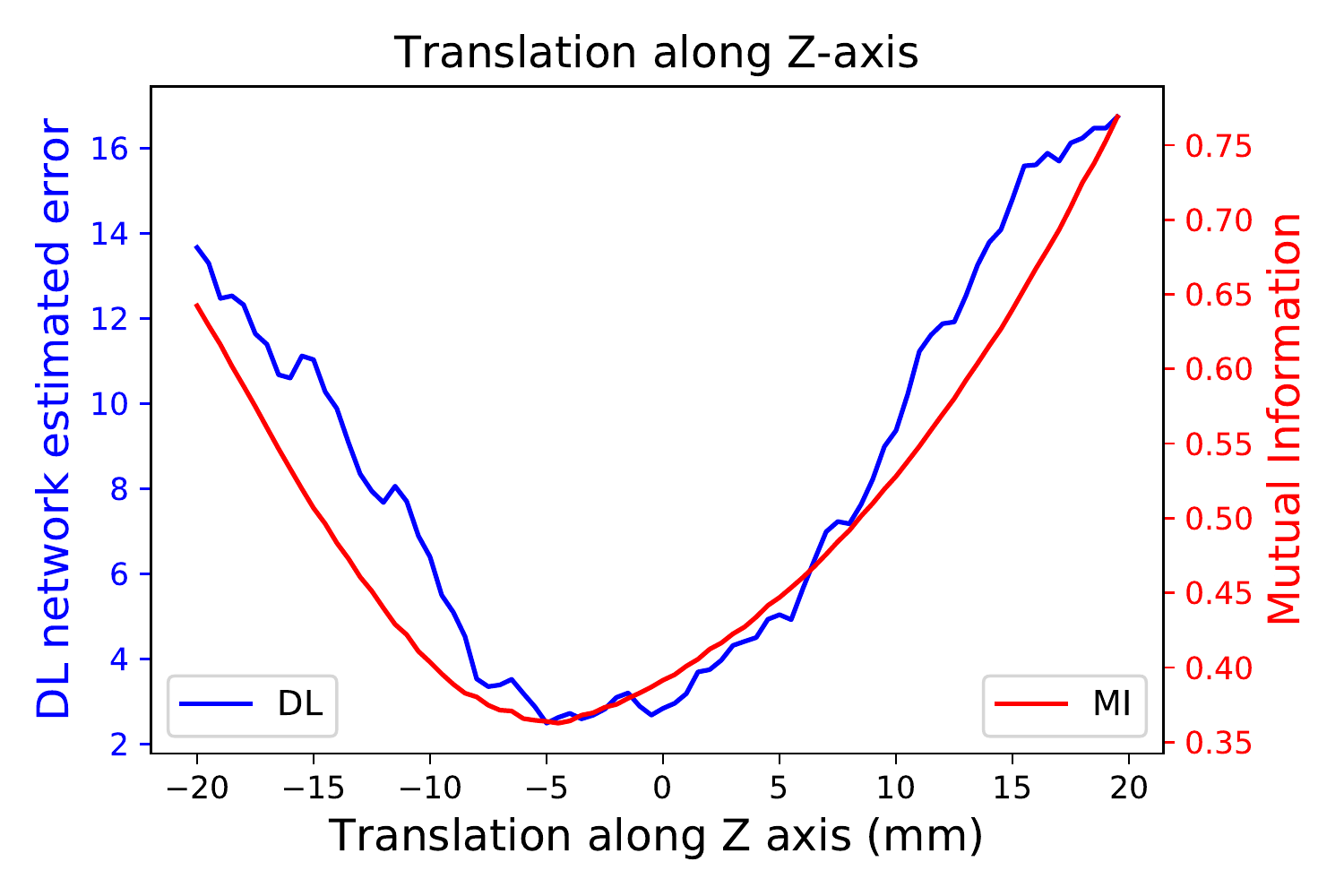}}
  \subfloat[]{\includegraphics[width=0.5\textwidth]{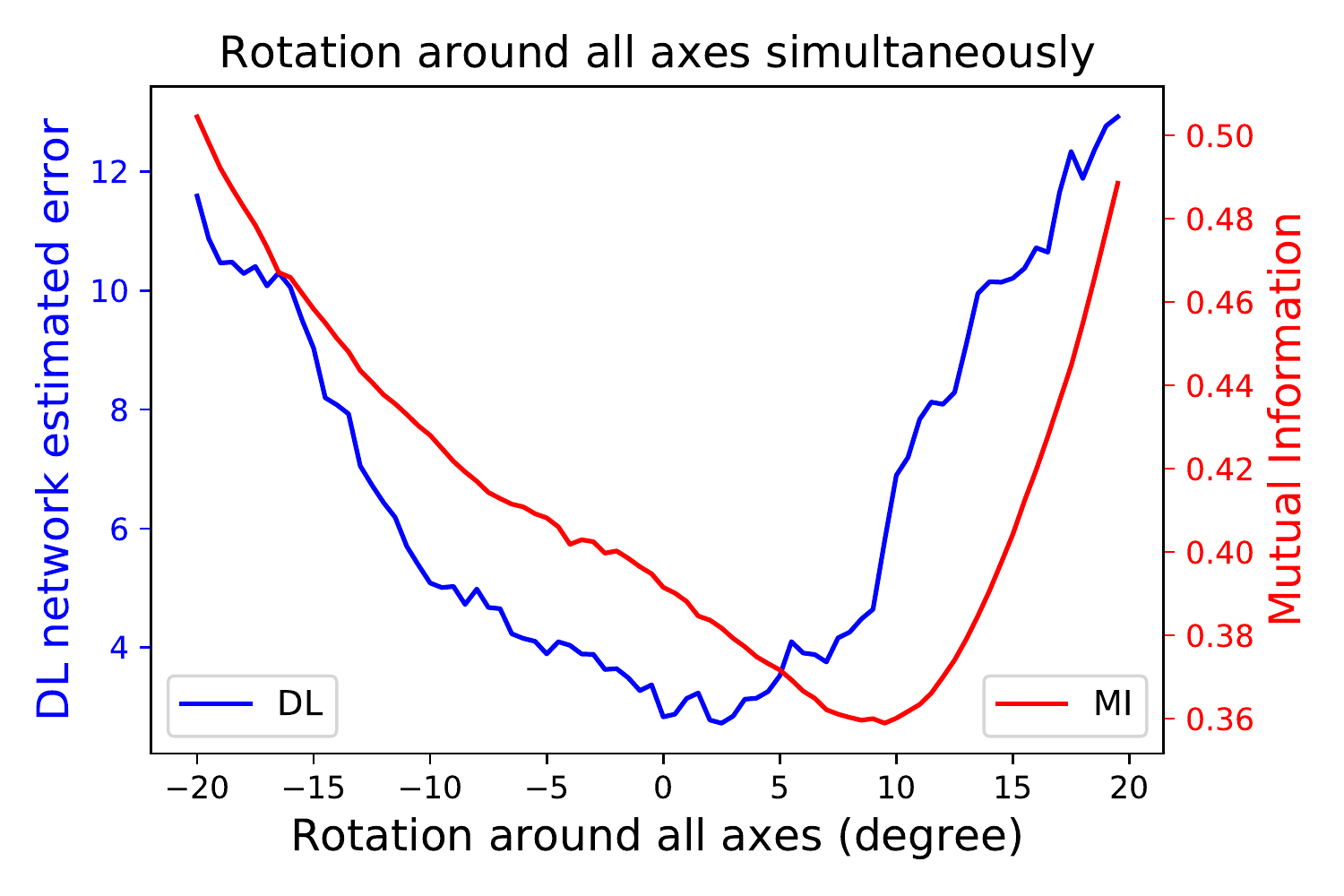}}

\caption{Curves that show the metric values that are outputted by the learned deep similarity metric using the multipass approach (blue) and mutual information (red) following (a) translation along the z-axis and (b) simultaneous rotations around all axes.}
\label{fig:metric_curves}
\end{figure}

In order to demonstrate the learned metric's ability to predict the target registration error, we use known geometric perturbations from ground truth and compare the outputs obtained using mutual information and the learned similarity metric. The curves of metric values versus the ground truth perturbation values are shown in Fig.~\ref{fig:metric_curves}. The curves show that the global optimum of the learned similarity metric is significantly closer to the ground truth, expert alignment than that of mutual information. However, the lack of smoothness/convexity of the learned metric is also apparent. This means that a robust optimization method is necessary. Without such a method, premature convergence to one of the local minimum could happen, which would significantly compromise the accuracy of the registration and fail to adequately utilize the superior learned metric.

\section{Experiments}
\label{sec:experiments}

\subsection{Materials}

A total 679 sets of data from the National Institute of Health (NIH) have been used for experiments. The data were all acquired from different MR-TRUS fusion-guided prostate cancer biopsy procedures, under an IRB-approved clinical trial after written informed consent of the participants/patients was obtained. Each set contains a T2-weighted MR volume and a reconstructed 3D ultrasound volume. The MR volume has 512$\times$512$\times$26 voxels with a resolution of 0.3mm$\times$0.3mm$\times$3mm, acquired with endorectal coil. In this application, the TRUS volumes were acquired using an end-fire probe sweeping through the prostate from base to apex in axial view. The ultrasound volumes have varying sizes and resolutions that are determined by the ultrasound scanning parameters associated with the reconstruction algorithm. The data are split into training, validation, and test sets consisting of 539, 70, and 70 cases, respectively.

\subsection{Registration Results}

\begin{figure}[thb]
	\centering
	\includegraphics[width=\textwidth]{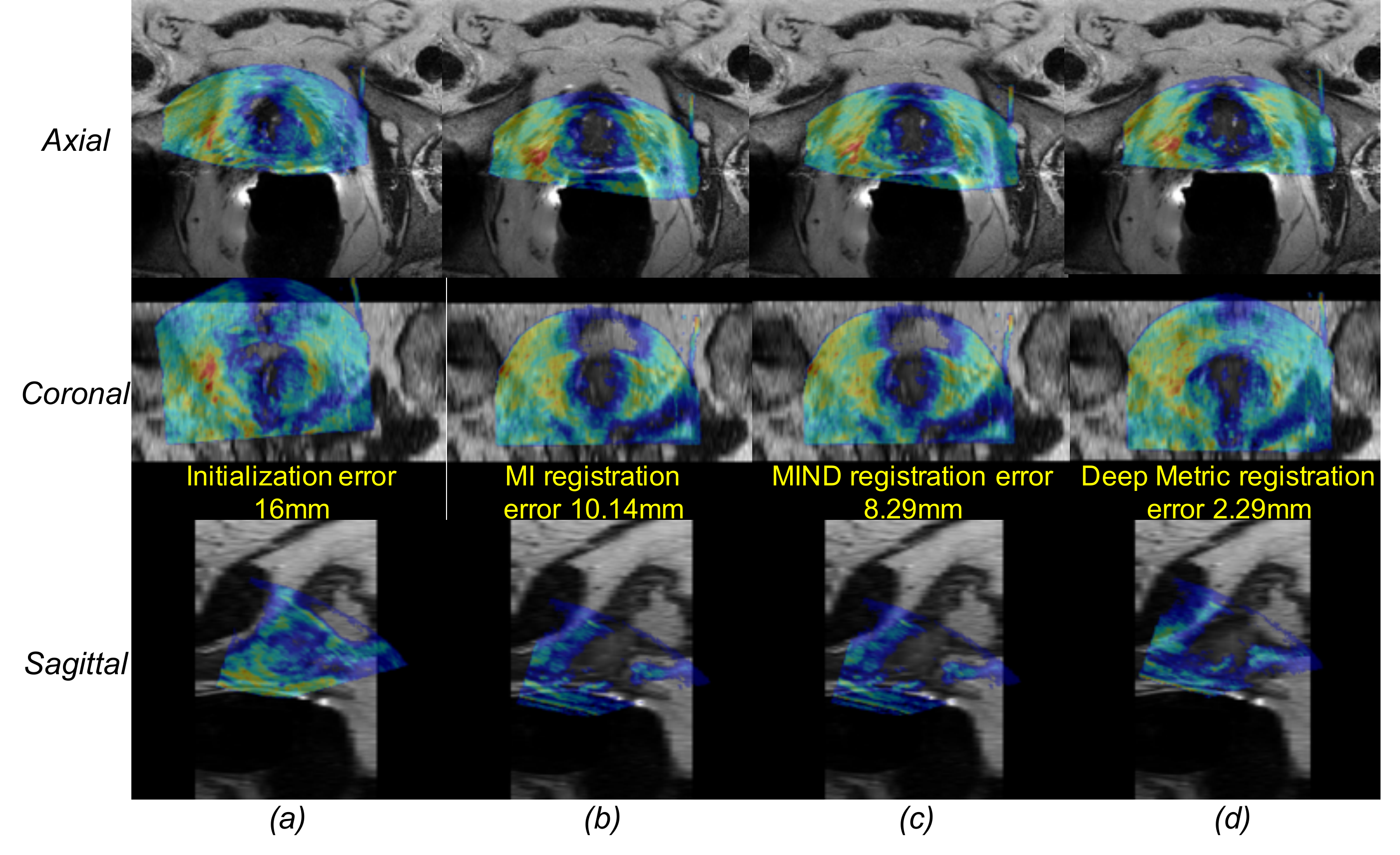}
	\caption{Axial, coronal, and sagittal views of example registration results: (a) Initial alignment; (b) Registration performed by optimizing the mutual information (MI) using DINO; (c) Registration performed by optimizing the MIND similarity using DINO; (d) Registration performed by optimizing the learned metric using DINO.}
	\label{fig:result_images}       
\end{figure}

It is necessary to determine the efficacy of the learned metric, to determine the efficacy of DINO, to demonstrate the utility of the multi-pass approach, and to assess the capture range associated with the proposed approach. The efficacy of the learned metric was determined by comparing its registration accuracy using DINO to that of both mutual information \cite{wells1996multi} and the sum of squared differences of the MIND descriptors of each image \cite{HeinrichMIND} using DINO. In order to demonstrate DINO's robustness to poor initialization, initializations of 8mm and 16mm are used for each method evaluation. The efficacy of the multi-pass approach is demonstrated by contrasting its results with those associated with the single-pass approach. A visualization of the registration results obtained using both mutual information and the learned metric as the similarity metric and DINO as the optimizer are given in Fig.~\ref{fig:result_images}.

%
\begin{table}
	\caption{Comparison of the TREs obtained using several different similarity metrics and optimization strategies. Note that (sp) refers to the single pass approach and (mp) refers to the multi-pass approach. All the numbers are in millimeter (mm).}
	\label{tab:opt_comparison}       
	\centering
	\begin{tabular}{l|l|c|lc}
		\hline
		Similarity Metric & Optimizer & Initial & Final mean$\pm$std & [min, max]\\
		\hline
		Mutual Information	& DINO	& \multirow{6}{*}{8mm}	&	8.96 $\pm$ 1.28 & [5.50, 13.45]\\
        SSD MIND			& DINO	& 	&	6.42 $\pm$ 2.86 & [1.75,  10.64]\\
        Deep Metric (sp)	& DINO 	& 	& 3.97 $\pm$ 1.67 & [0.77,  8.51]\\
		Deep Metric (mp)	& BFGS 	& 	& 7.31 $\pm$ 0.61 & [6.63,  9.11]\\
		Deep Metric (mp)	& Powell 	&  &  6.11 $\pm$ 4.62 & [1.89,  12.98]\\
		Deep Metric	(mp)	& DINO 	&  & \textbf{3.82 $\pm$ 1.63} & [0.65,  8.80]\\
		\hline
		Mutual Information & DINO	& \multirow{6}{*}{16mm}	&	10.07 $\pm$ 1.40 & [8.82,  14.09]\\
        SSD MIND	    & DINO	& 	&	6.62 $\pm$ 2.96 & [1.58, 13.63]\\
		Deep Metric (sp)	& DINO 	& 	& 4.21 $\pm$ 1.64 & [1.08, 8.68]\\
        Deep Metric (mp)	& BFGS 	& 	& 14.27 $\pm$ 0.61 & [13.11, 15.25]\\
		Deep Metric (mp)	& Powell & 	& 11.05 $\pm$ 5.32 & [2.11, 24.09]\\
		Deep Metric (mp)	& DINO 	& 	& \textbf{3.94 $\pm$ 1.47} & [1.35,  9.37]\\
		\hline
	\end{tabular}
\end{table}

In addition to evaluating its capture range, DINO's ability to produce accurate registrations using the learned metric is also be demonstrated. DINO's performance is contrasted with that of BFGS \cite{BFGS} and Powell's method \cite{MJDPowell} without any prior solution space exploration. The results displayed in Table~\ref{tab:opt_comparison} show that using the learned similarity metric results in more accurate registration that is robust to the registration error associated with the initialization. Furthermore, the performance of DINO is clearly superior to that of the other two optimization methods regardless of the initial registration error. Therefore, both the learned metric and DINO are needed in order to robustly obtain quality registrations. The mean and standard deviation are calculated using the TREs that result from performing 70 registrations, one for each test case.

\subsection{Implications and Discussions}

Our results indicate that the similarity metric that is learned by the CNN captures the TRE better than the mutual information and MIND-based  metrics that are commonly used for multimodal registration. As a result, the TREs obtained using our registration framework are smaller than those obtained by the other two aforementioned intensity-based MR-TRUS registration approaches. The optimization strategy that we use (DINO) is also able to handle the nonconvex, nonsmooth morphology of the similarity metric. Further, the results show that the TRE obtained using our registration pipeline is relatively invariant to the quality of the initial alignment. The use of a distribution-based optimization strategy in the first stage provided the algorithm with a large capture range. This is an important result as small capture ranges limit many registration applications. 

Although the proposed approach is robust to poor initializations and outperforms the previous state-of-the-art MIND based registration with improved registration accuracy, extra local adjustment needs to be performed by physicians when targeting small lesions. Because inexperienced physicians may struggle with 3D MR-TRUS registration, it is important to design an automatic registration algorithm that is invariant to potentially poor initializations. Furthermore, our software allows a user to override the registration and saves the solution produced by either manual correction or fully automatic registration. It provides instant feedback to the user that assesses the quality of any alteration that is made using a simple forward pass through our CNN. A physician has the option to save any initialization or registration for later comparison, manually perform MR-TRUS registration that is assessed by the learned metric, and/or use DINO to automatically perform the registration by optimizing the learned metric. The variety of options that our software provides and its simple user interface are conducive to faster clinical adoption. 

It should be noted that this work used a rigid transformation model instead of a deformable one, which inherently limits the registration accuracy due to the deformation of the prostate that occurs between image acquisitions. Rigid registration could be sufficient for our application most of the time as the deformation is rather limited. However, deep similarity based deformable registration will be investigated in the future for more comprehensive evaluation. 

\section{Conclusion}
\label{sec:conclusions}

The search for a metric of registration accuracy is driven by an evolving clinical need for more accurate registration of MR-TRUS volumes. Although fusion biopsy is dependent upon registration, it has been very challenging to achieve accurate registration due to the large appearance difference between TRUS and MR images. As MR gets more and more sensitive in the detection of small lesions, sampling those small lesions becomes increasingly dependent upon the targeting accuracy of the operator. A lack of reproducible metrics and standardization tools leads to a reduction in clinical impact, less uniform practice, and may limit the success of this novel technology in clinical practice. 

In this paper, the developed similarity metric is learned using a custom convolutional neural network and demonstrates very promising performance using a composite optimization scheme to perform the registration. In our future work, we hope to explore methods to enforce desirable similarity metric morphology, investigate variants of our neural network training strategy, further speed up the registration, extend this approach to the deformable registration case, and encourage more computationally efficient and thorough exploration of the solution space in future works.

\begin{acknowledgements}
We would also like to thank NVIDIA Corporation for the donation of the Titan Xp GPU used for this research (PY).
\end{acknowledgements}

\section*{Compliance with ethical standards}

\absdiv{Conflict of interest} NIH and Philips/In Vivo have a Cooperative Research and Development Agreement. NIH and Philips share intellectual property in the field and one author receive royalties for licensed patents (BW). PY was a salaried employee of Philips Research at the time some of the research was performed.

\absdiv{Ethical approval} All procedures performed in studies involving human participants were in accordance with the ethical standards of the institutional and/or national research committee of where the studies were
conducted.

\absdiv{Informed consent} Informed consent was obtained from all individual participants included in the study.


\begin{thebibliography}{26}
	\providecommand{\natexlab}[1]{#1}
	\providecommand{\url}[1]{{#1}}
	\providecommand{\urlprefix}{URL }
	\expandafter\ifx\csname urlstyle\endcsname\relax
	\providecommand{\doi}[1]{DOI~\discretionary{}{}{}#1}\else
	\providecommand{\doi}{DOI~\discretionary{}{}{}\begingroup
		\urlstyle{rm}\Url}\fi
	\providecommand{\eprint}[2][]{\url{#2}}
	
	\bibitem[{Calio et~al(2017)Calio, Sidana, Sugano, Gaur, Jain, Maruf, Xu, Yan,
		Kruecker, Merino, Choyke, Turkbey, Wood, and Pinto}]{calio_changes_2017}
	Calio B, Sidana A, Sugano D, Gaur S, Jain A, Maruf M, Xu S, Yan P, Kruecker J,
	Merino M, Choyke P, Turkbey B, Wood B, Pinto P (2017) Changes in prostate
	cancer detection rate of {MRI}-{TRUS} fusion vs systematic biopsy over time:
	evidence of a learning curve. Prostate Cancer And Prostatic Diseases 20:436
	
	\bibitem[{Cheng et~al(2016)Cheng, Zhang, and Zheng}]{cheng_deep_2016}
	Cheng X, Zhang L, Zheng Y (2016) Deep similarity learning for multimodal
	medical images. Computer Methods in Biomechanics and Biomedical Engineering:
	Imaging \& Visualization pp 1--5
	
	\bibitem[{Chollet(2015)}]{keras}
	Chollet F (2015) Keras. \url{https://github.com/fchollet/keras}
	
	\bibitem[{Fedorov et~al(2015)Fedorov, Khallaghi, S{\'a}nchez, Lasso, Fels,
		Tuncali, Sugar, Kapur, Zhang, Wells, Nguyen, Abolmaesumi, and
		Tempany}]{Fedorov2015}
	Fedorov A, Khallaghi S, S{\'a}nchez CA, Lasso A, Fels S, Tuncali K, Sugar EN,
	Kapur T, Zhang C, Wells W, Nguyen PL, Abolmaesumi P, Tempany C (2015)
	Open-source image registration for mri--trus fusion-guided prostate
	interventions. International Journal of Computer Assisted Radiology and
	Surgery 10(6):925--934
	
	\bibitem[{Fletcher and Powell(1963)}]{MJDPowell}
	Fletcher R, Powell MJD (1963) A {Rapidly} {Convergent} {Descent} {Method} for
	{Minimization}. The Computer Journal 6(2):163--168,
	\doi{10.1093/comjnl/6.2.163},
	\urlprefix\url{http://dx.doi.org/10.1093/comjnl/6.2.163}
	
	\bibitem[{Fuerst et~al(2014)Fuerst, Wein, Müller, and
		Navab}]{fuerst_automatic_2014}
	Fuerst B, Wein W, Müller M, Navab N (2014) Automatic ultrasound-{MRI}
	registration for neurosurgery using the 2d and 3d {LC}(2) {Metric}. Medical
	Image Analysis 18(8):1312--1319, \doi{10.1016/j.media.2014.04.008}
	
	\bibitem[{Heinrich et~al(2012)Heinrich, Jenkinson, Bhushan, Matin, Gleeson,
		Brady, and Schnabel}]{HeinrichMIND}
	Heinrich MP, Jenkinson M, Bhushan M, Matin T, Gleeson FV, Brady SM, Schnabel JA
	(2012) {MIND}: Modality independent neighbourhood descriptor for multi-modal
	deformable registration. Medical Image Analysis 16(7):1423 -- 1435
	
	\bibitem[{Jones et~al(2001--)Jones, Oliphant, and Peterson}]{scipy}
	Jones E, Oliphant T, Peterson P (2001--) {SciPy}: Open source scientific tools for {Python}. \urlprefix\url{http://www.scipy.org/}, [Online; accessed
	2018-07-30]
	
	\bibitem[{Khallaghi et~al(2015)Khallaghi, Sánchez, Rasoulian, Sun, Imani,
		Khojaste, Goksel, Romagnoli, Abdi, Chang, Mousavi, Fenster, Ward, Fels, and
		Abolmaesumi}]{khallaghi_biomechanically_2015}
	Khallaghi S, Sánchez CA, Rasoulian A, Sun Y, Imani F, Khojaste A, Goksel O,
	Romagnoli C, Abdi H, Chang S, Mousavi P, Fenster A, Ward A, Fels S,
	Abolmaesumi P (2015) Biomechanically {Constrained} {Surface} {Registration}:
	{Application} to {MR}-{TRUS} {Fusion} for {Prostate} {Interventions}. IEEE
	Transactions on Medical Imaging 34(11):2404--2414
	
	\bibitem[{Krizhevsky et~al(2012)Krizhevsky, Sutskever, and
		Hinton}]{krizhevsky_imagenet_2012}
	Krizhevsky A, Sutskever I, Hinton GE (2012) {ImageNet} {Classification} with
	{Deep} {Convolutional} {Neural} {Networks}. In: Proceedings of the 25th
	{International} {Conference} on {Neural} {Information} {Processing} {Systems}
	- {Volume} 1, Curran Associates Inc., USA, {NIPS}'12, pp 1097--1105
	
	\bibitem[{LeCun et~al(2015)LeCun, Bengio, and Hinton}]{lecun_deep_2015}
	LeCun Y, Bengio Y, Hinton G (2015) Deep learning. Nature 521(7553):436
	
	\bibitem[{Maes et~al(1997)Maes, Collignon, Vandermeulen, Marchal, and
		Suetens}]{maes1997multimodality}
	Maes F, Collignon A, Vandermeulen D, Marchal G, Suetens P (1997) Multimodality
	image registration by maximization of mutual information. IEEE transactions
	on Medical Imaging 16(2):187--198
	
	\bibitem[{Pinto et~al(2011)Pinto, Chung, Rastinehad, Baccala, Kruecker,
		Benjamin, Xu, Yan, Kadoury, Chua, Locklin, Turkbey, Shih, Gates, Buckner,
		Bratslavsky, Linehan, Glossop, Choyke, and Wood}]{pinto_magnetic_2011}
	Pinto PA, Chung PH, Rastinehad AR, Baccala AA, Kruecker J, Benjamin CJ, Xu S,
	Yan P, Kadoury S, Chua C, Locklin JK, Turkbey B, Shih JH, Gates SP, Buckner
	C, Bratslavsky G, Linehan WM, Glossop ND, Choyke PL, Wood BJ (2011) Magnetic
	resonance imaging/ultrasound fusion guided prostate biopsy improves cancer
	detection following transrectal ultrasound biopsy and correlates with
	multiparametric magnetic resonance imaging. The Journal of Urology
	186(4):1281--1285
	
	\bibitem[{Poulin et~al(2018)Poulin, Boudam, Pinter, Kadoury, Lasso, Fichtinger,
		and Ménard}]{poulin_validation_2018}
	Poulin E, Boudam K, Pinter C, Kadoury S, Lasso A, Fichtinger G, Ménard C
	(2018) Validation of {MRI} to {TRUS} registration for high-dose-rate prostate
	brachytherapy. Brachytherapy 17(2):283--290,
	\doi{10.1016/j.brachy.2017.11.018}
	
	\bibitem[{Reena~Benjamin and Jayasree(2018)}]{reena_benjamin_improved_2018}
	Reena~Benjamin J, Jayasree T (2018) Improved medical image fusion based on
	cascaded {PCA} and shift invariant wavelet transforms. International Journal
	of Computer Assisted Radiology and Surgery 13(2):229--240
	
	\bibitem[{Shanno(1970)}]{BFGS}
	Shanno DF (1970) Conditioning of {Quasi}-{Newton} {Methods} for {Function}
	{Minimization}. Mathematics of Computation 24(111):647--656,
	\doi{10.2307/2004840}, \urlprefix\url{http://www.jstor.org/stable/2004840}
	
	\bibitem[{Siddiqui et~al(2015)Siddiqui, Rais-Bahrami, Turkbey, George, Rothwax,
		Shakir, Okoro, Raskolnikov, Parnes, Linehan, Merino, Simon, Choyke, Wood, and
		Pinto}]{siddiqui_comparison_2015}
	Siddiqui MM, Rais-Bahrami S, Turkbey B, George AK, Rothwax J, Shakir N, Okoro
	C, Raskolnikov D, Parnes HL, Linehan WM, Merino MJ, Simon RM, Choyke PL, Wood
	BJ, Pinto PA (2015) Comparison of {MR}/ultrasound fusion-guided biopsy with
	ultrasound-guided biopsy for the diagnosis of prostate cancer. JAMA
	313(4):390--397
	
	\bibitem[{Siegel et~al(2018)Siegel, Miller, and Jemal}]{siegel_cancer_nodate}
	Siegel RL, Miller KD, Jemal A (2018) Cancer statistics. CA: A Cancer Journal
	for Clinicians
	
	\bibitem[{Simonovsky et~al(2016)Simonovsky, Gutiérrez-Becker, Mateus, Navab,
		and Komodakis}]{simonovsky_deep_2016}
	Simonovsky M, Gutiérrez-Becker B, Mateus D, Navab N, Komodakis N (2016) A
	{Deep} {Metric} for {Multimodal} {Registration}. In: Medical {Image}
	{Computing} and {Computer}-{Assisted} {Intervention} -- {MICCAI}, pp 10--18
	
	\bibitem[{Sparks et~al(2013)Sparks, Bloch, Feleppa, Barratt, and
		Madabhushi}]{sparks_fully_2013}
	Sparks R, Bloch BN, Feleppa E, Barratt D, Madabhushi A (2013) Fully {Automated}
	{Prostate} {Magnetic} {Resonance} {Imaging} and {Transrectal} {Ultrasound}
	{Fusion} via a {Probabilistic} {Registration} {Metric}. In: SPIE Medical
	Imaging, vol 8671
	
	\bibitem[{Storn and Price(1997)}]{Storn1997}
	Storn R, Price K (1997) Differential evolution -- a simple and efficient
	heuristic for global optimization over continuous spaces. Journal of Global
	Optimization 11(4):341--359
	
	\bibitem[{Sun et~al(2013)Sun, Yuan, Rajchl, Qiu, Romagnoli, and
		Fenster}]{sun_efficient_2013}
	Sun Y, Yuan J, Rajchl M, Qiu W, Romagnoli C, Fenster A (2013) Efficient
	{Convex} {Optimization} {Approach} to 3d {Non}-rigid {MR}-{TRUS}
	{Registration}. In: Mori K, Sakuma I, Sato Y, Barillot C, Navab N (eds)
	Medical {Image} {Computing} and {Computer}-{Assisted} {Intervention} –
	{MICCAI} 2013, Springer Berlin Heidelberg, pp 195--202
	
	\bibitem[{Sun et~al(2015)Sun, Yuan, Qiu, Rajchl, Romagnoli, and
		Fenster}]{sun_three-dimensional_2015}
	Sun Y, Yuan J, Qiu W, Rajchl M, Romagnoli C, Fenster A (2015)
	Three-{Dimensional} {Nonrigid} {MR}-{TRUS} {Registration} {Using} {Dual}
	{Optimization}. IEEE Transactions on Medical Imaging 34(5):1085--1095
	
	\bibitem[{Wells et~al(1996)Wells, Viola, Atsumi, Nakajima, and
		Kikinis}]{wells1996multi}
	Wells WM, Viola P, Atsumi H, Nakajima S, Kikinis R (1996) Multi-modal volume
	registration by maximization of mutual information. Medical image analysis
	1(1):35--51
	
	\bibitem[{Zagoruyko and Komodakis(2015)}]{Zagoruyko_Komodakis2015}
	Zagoruyko S, Komodakis N (2015) Learning to compare image patches via
	convolutional neural networks. In: IEEE Conf. Computer Vision and Pattern
	Recognition (CVPR), pp 4353--4361
	
	\bibitem[{Zettinig et~al(2015)Zettinig, Shah, Hennersperger, Eiber, Kroll,
		Kübler, Maurer, Milletarì, Rackerseder, Schulte~zu Berge, Storz, Frisch,
		and Navab}]{zettinig_multimodal_2015}
	Zettinig O, Shah A, Hennersperger C, Eiber M, Kroll C, Kübler H, Maurer T,
	Milletarì F, Rackerseder J, Schulte~zu Berge C, Storz E, Frisch B, Navab N
	(2015) Multimodal image-guided prostate fusion biopsy based on automatic
	deformable registration. International Journal of Computer Assisted Radiology
	and Surgery 10(12):1997--2007
	
\end{thebibliography}

\end{document}